\pdfoutput=1

\documentclass[11pt]{article}

\usepackage[]{ACL2023}

\usepackage{times}
\usepackage{latexsym}

\usepackage[T1]{fontenc}

\usepackage[utf8]{inputenc}
\usepackage{amsmath}
\usepackage{amsfonts}
\usepackage{amssymb}
\usepackage{microtype}
\usepackage{booktabs}
\usepackage{multirow}

\usepackage{inconsolata}

\usepackage{graphicx}

\usepackage[utf8]{inputenc}

\title{Improving Factuality of Abstractive Summarization \\ via Contrastive Reward Learning}

\author{
I-Chun Chern\textsuperscript{\rm{1}} \quad Zhiruo Wang\textsuperscript{\rm{1}} \quad Sanjan Das\textsuperscript{\rm{1}} \quad Bhavuk Sharma\textsuperscript{\rm{1}} \quad  \\
\textbf{Pengfei Liu}\textsuperscript{\rm{2}} \quad \textbf{Graham Neubig}\textsuperscript{\rm{1}} \\
  \textsuperscript{1} Carnegie Mellon University \\
  \textsuperscript{2}  Shanghai Jiao Tong University\\
  {\tt \{ichern, zhiruow, sanjand, bhavuks, gneubig\}@cs.cmu.edu} \\
  {\tt stefanpengfei@gmail.com} \\
}

\begin{document}
\maketitle
\begin{abstract}
Modern abstractive summarization models often generate summaries that contain hallucinated or contradictory information. In this paper, we propose a simple but effective contrastive learning framework that incorporates recent developments in reward learning and factuality metrics. Empirical studies demonstrate that the proposed framework enables summarization models to learn from feedback of factuality metrics using contrastive reward learning, leading to more factual summaries by human evaluations. This suggests that further advances in learning and evaluation algorithms can feed directly into providing more factual summaries. Code and human evaluation results will be publicly available at \url{https://github.com/EthanC111/factuality_summarization}.
\end{abstract}

\section{Introduction}
One major challenge in current abstractive summarization models is how to generate more factual summaries that are consistent with the source text~\citep{li2022faithfulness}. Various approaches have been proposed to address this challenge, including augmenting the model input \citep{dou-etal-2021-gsum}, performing post-processing \citep{dong-etal-2020-multi, cao-etal-2020-factual}, and modifying the learning algorithms \citep{cao-wang-2021-cliff, liu2021co2sum}.
In particular, learning-based methods possess the advantage of not requiring modification to the existing model architecture or the addition of new modules.

In the meantime, with the growing interest in aligning learning objectives with evaluation criteria of interest, utilizing feedback of automatic evaluation metrics \citep{liu-etal-2022-brio} or human preferences \citep{stiennon2020learning} as rewards for fine-tuning abstractive summarization models has gained substantial attention. These methods learn to optimize rewards using techniques such as reinforcement-learning (RL) \citep{stiennon2020learning}, minimum risk training (MRT) \cite{shen-etal-2016-minimum, wieting-etal-2019-beyond}, and contrastive reward learning (CRL) \cite{liu-liu-2021-simcls, liu-etal-2022-brio}.

\begin{figure}
  \centering
  \includegraphics[width=0.5\textwidth]{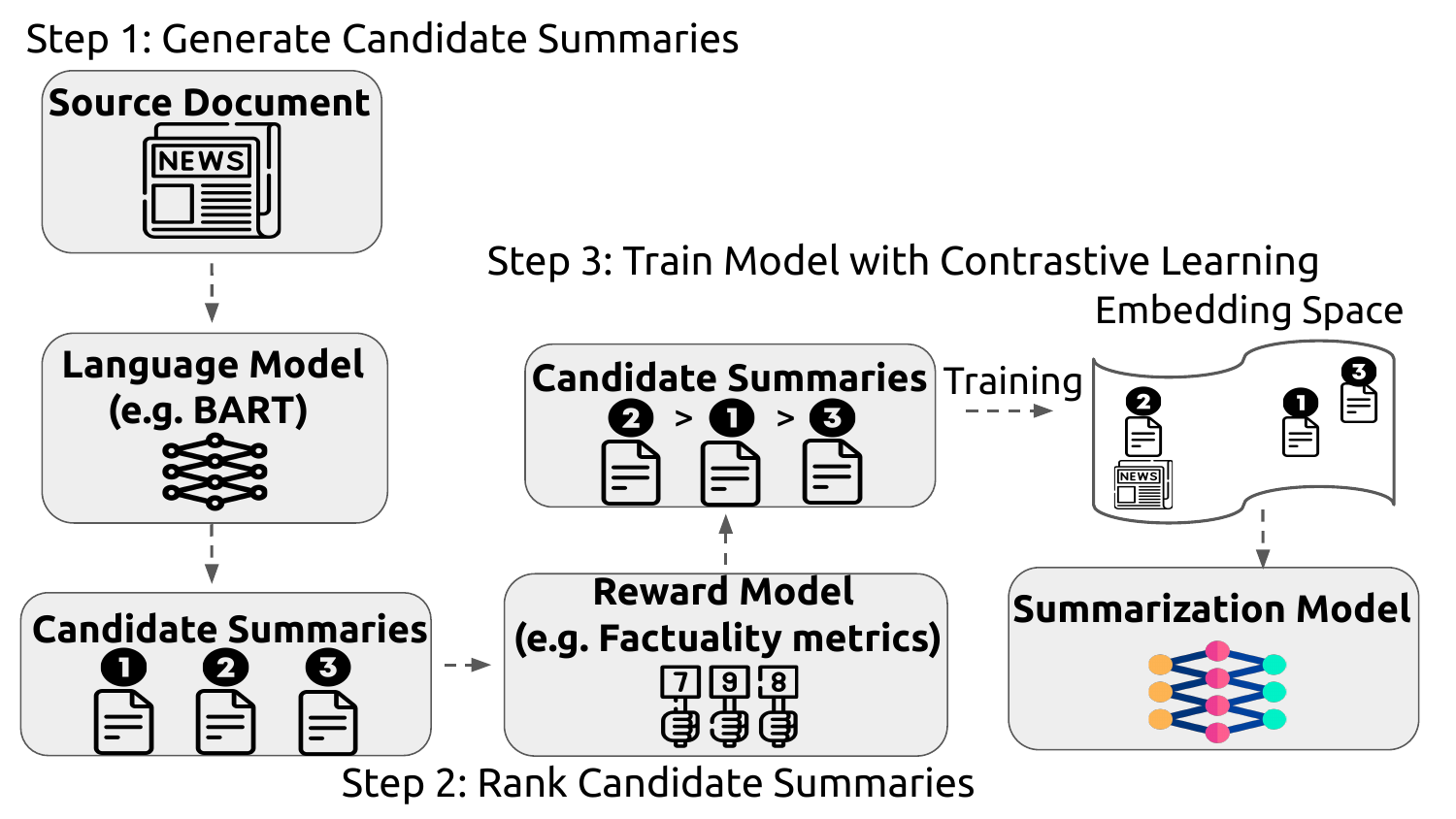}
  \caption{An illustration of our learning framework.}
  \label{fig:yourlabel }
\end{figure}

Given the benefits of learning-based methods in improving factuality of abstractive summarization, and recent advancements in factuality metrics for detecting factual inconsistencies in generated summaries, it is of interest to apply reward learning to enforce models to learn from feedback of factuality metrics to improve the factuality of abstractive summarization models. We aim to investigate the following questions in this paper - \textbf{Q1:} Can contrastive reward learning effectively utilize existing factuality metrics to improve the factuality of abstractive summarization models? \textbf{Q2:} Can the improvement in factuality be reflected in human evaluation studies?

In this paper, we propose a contrastive reward learning framework that enables abstractive summarization models to directly learn from feedback of factuality metrics in a sample-efficient manner. In contrast to other contrastive learning frameworks \citep{cao-wang-2021-cliff, liu2021co2sum}, our proposed framework does not rely on the complex construction of negative samples. Instead, similar to \citep{liu-etal-2022-brio}, all candidate summaries used for contrastive learning are generated from pretrained sequence-to-sequence models \citep{lewis-etal-2020-bart, zhang2020pegasus} using diverse beam search \citep{vijayakumar2018diverse}. Additionally, our framework also incorporates the use of quality metrics to provide more fine-grained information on the ranking (positive / negative) of candidate summaries. Specifically, we investigate learning from the rewards of two factuality metrics: BARTScore  \cite{yuan2021bartscore} 
and DAE \cite{goyal-durrett-2021-annotating}. Through automatic and human evaluation studies,  we demonstrate that our framework enables summarization models to generate significantly more factual summaries.

\section{Contrastive Learning from Factuality Rewards}

\subsection{Contrastive Learning for Abstractive Summarization}

\paragraph{Abstractive Summarization}
Given a source document $D$, the summarization model learns a generative model $g_\theta$, that converts the source document $D$ into a summary $S$:

\begin{equation}
    S = g_\theta(D)
\label{eq:abs_sum}
\end{equation}

\paragraph{MLE Loss}
Given a training sample pair $\{D,S^{r}\}$ consists of source document $D$ and reference summary $S^r$ (note that $S^{r}$ consists of $L$ tokens, $S^{r} = \{s^{r}_{1}, \cdots, s^{r}_{j}, \cdots, s^{r}_{L}\}$), the MLE loss $\mathcal{L}_{\text{mle}}$ aims to maximize the likelihood of reference summary $S^{r}$ given the source document $D$: 
\begin{equation}
\mathcal{L}_{\text{mle}} = \log p_{g_\theta}(S^r | D) =  \sum_{j = 1}^{L}\log p_{g_\theta}(s^r_j | D, s^{r}_{<j})
\label{eq:mle_loss}
\end{equation}
where $s^{r}_{<j} = \{s^r_0, \cdots, s^r_{j-1}\}$ and $s^r_0$ is a pre-defined start token.

Despite its effectiveness in enforcing generated summaries to align with the reference summaries, the MLE loss is not aware of the \textit{quality} (evaluated by some quality metric $M$) of the generated summaries. To address this issue, we introduce a contrastive loss \cite{liu-etal-2022-brio}.

\paragraph{Contrastive Loss}
Given a training sample pair $\{D,S^{r}\}$, and that $S_i, S_j$ are candidate summaries generated from a pre-trained model given $D$, and that $M(S_i) > M(S_j)$ $\forall i,j,i < j$ \footnote{
$M$ could be reference-free (e.g., BARTScore, DAE) or reference-based (e.g., ROUGE) metric. If $M$ is a reference-free metric, then $M(S_i) = M(S_i, D)$ ; if $M$ is a reference-based metric, then $M(S_i) = M(S_i, S^{r})$}, the contrastive loss is defined as:

\begin{equation}
\mathcal{L}_{ctr} = \sum_i \sum_{j > i}\max (0, f(S_j) - f(S_i) + \lambda_{ij})
\label{eq:ctr_loss}
\end{equation}
Note that $\lambda_{ij} = (j - i) \times \lambda$ is the rank difference between two candidates times a constant $\lambda$ (usually set as $1$) \footnote{The magnitude of contrastive loss can be directly regulated through the weight of contrastive loss $\gamma$, so we simply set $\lambda$ equal to 1.} and that $f(S)$ is the length-normalized estimated log-probability given by:

\begin{equation}
f(S) = \frac{\sum_{t=1}^{l}\log p_{g_\theta}(s_t|D, S_{<t})}{|S|^\alpha}
\end{equation}
where $\alpha$ is a constant.

Intuitively, the contrastive loss penalizes any discoordination between the length-normalized estimated log-probability and the quality metric evaluation (i.e., when $f(S_j) > f(S_i)$ but $M(S_i) > M(S_j)$). The quality metric $M$ could be any evaluation criteria, including automatic evaluation metrics \cite{lin-2004-rouge, yuan2021bartscore, goyal-durrett-2021-annotating}, or human preferences \cite{ouyang2022training}.

\paragraph{Combined Loss}
The combined loss used for fine-tuning is described by \autoref{eq:brio_loss}.

\begin{equation}
    \mathcal{L}_{com}=\mathcal{L}_{\text{mle}}+\gamma \mathcal{L}_{ctr}
\label{eq:brio_loss}
\end{equation}

where $\mathcal{L}_{\text{mle}}$ is the MLE loss given in \autoref{eq:mle_loss}, $\mathcal{L}_{ctr}$ is the contrastive loss given in \autoref{eq:ctr_loss}, and $\gamma$ is the weight of contrastive loss. Summarization models fine-tuned with $\mathcal{L}_{com}$ is referred as CRL-COM.

\subsection{Reward from Factuality Metrics}
We use two factuality metrics as quality metrics $M$ for use in the contrastive loss described in \autoref{eq:ctr_loss}.

\paragraph{BARTScore} \cite{yuan2021bartscore}'s factuality score is calculated as the log-likelihood of the summary given the source calculated from a reference-free version of BARTScore.

\paragraph{DAE} \cite{goyal-durrett-2021-annotating} is calculated as the softmax output of the least-factual dependency-arc inside the sentences in the summary.

These two metrics were chosen for relative computational efficiency, as they are evaluated many times in the training process.
\footnote{
In contrast, QA-based factuality metrics are computationally inefficient \citep{laban-etal-2022-summac}. As a result, they are less feasible for use in reward-learning settings.
}

\section{Experiments}

\subsection{Experimental Setup} 
Driven by the two research questions presented in the introduction, we train two kinds of factuality-driven summarization models, namely CRL-COM (B) and CRL-COM (D), trained from contrastive reward learning using BARTScore and DAE as quality metrics, respectively. A baseline summarization model CRL-COM (R) is also trained from contrastive reward learning using ROUGE as quality metric. Note that commonly used n-gram based metrics, including ROUGE \citep{lin-2004-rouge}, have been shown to have a low correlation with human evaluations, particularly on factuality perspective \citep{falke2019ranking, durmus-etal-2020-feqa}. Thus, we focus on evaluating the factuality of CRL-COM (B) and CRL-COM (D) compared to CRL-COM (R), with the hypothesis that CRL-COM (B) and CRL-COM (D) should be capable of generating more factual summaries compare to CRL-COM (R).

\paragraph{Datasets:}
We use two abstractive summarization datasets -- CNN/Daily Mail (CNNDM) dataset \citep{hermann2015teaching, nallapati2016abstractive} and the XSUM dataset \citep{narayan-etal-2018-dont}. CNNDM summaries tend to be more extractive and are composed of multi-sentence summaries, while XSUM summaries are more abstractive and are composed of single-sentence summaries.

\paragraph{Models:}
Following the setting outlined in \citep{liu-etal-2022-brio}, we fine-tuned a pre-trained BART model \citep{lewis-etal-2020-bart} on the CNNDM dataset and a pre-trained PEGASUS \citep{zhang2020pegasus} model on the XSUM dataset.

\paragraph{Implementation and Fine-tuning Details:}
The combined loss (with weight of the contrastive loss $\gamma = 100$) described in \autoref{eq:brio_loss} is used to fine-tune the pre-trained models. Following \citep{liu-etal-2022-brio} few-shot fine-tuning learning paradigm, we sampled 1000 training samples from each dataset for few-shot fine-tuning. A constant learning rate of $10^{-5}$ and $10^{-4}$ was applied to the fine-tuning process for the CNNDM and XSUM datasets, respectively, in order to facilitate fast convergence. For each dataset, we fine-tuned three models using three different quality metrics: ROUGE (R), BARTScore (B), and DAE (D), designated as CRL-COM (R), CRL-COM (B), and CRL-COM (D), respectively.
During validation, we employed the same quality metric used for fine-tuning for early stopping.

\paragraph{Automatic Evaluation} 
Each model is evaluated on three metrics: ROUGE (with variants ROUGE-1, ROUGE-2, ROUGE-L), BARTScore, and DAE.

\paragraph{Human Evaluation}
To objectively evaluate the factual consistencies of the generated summaries from each model, we randomly sampled 100 samples from CNNDM and 200 samples from XSUM for human evaluation. We assess each summary from three different perspectives: Factuality (FAC), Coherence (COH), and Relevance (REL), with a particular emphasis on factuality. The assessment follow similar guidelines as in \citep{liang2022holistic, fabbri2021summeval}. The evaluation guidelines provided to the annotators are listed in \autoref{tab:human_eval_guidelines}. An expert annotator is involved in the human evaluation studies.

\begin{table*}
\small
\centering
\resizebox{0.98\textwidth}{!}{
    \begin{tabular}{ll}
    \toprule
    \textbf{Perspective} & \multicolumn{1}{c}{\textbf{Guidelines}} \\
    \midrule
    \multirow{2}{*}{Factuality (FAC)} & {If all the information and claims inside the summary are included in the source article, } \\
    {} & {assign a binary score of 1 ; otherwise, assign a binary score of 0.} \\
    \midrule
    \multirow{2}{*}{Coherence (COH)} & {On a Likert scale of 1 (worst) to 5 (best), assign a score based on how well the} \\
    {} & {relevant information is coordinated and organized into a well-structured summary.} \\
    \midrule
    \multirow{2}{*}{Relevance (REL)} & {On a Likert scale of 1 (worst) to 5 (best), assign a score based on the extent to which } \\
    {} & {the summary includes only important information from the source article.} \\
    \bottomrule
    \end{tabular}
}
\caption{Guidelines for human evaluation studies}
\label{tab:human_eval_guidelines}
\end{table*}

\begin{table*}[ht]
\small
\centering
\resizebox{0.75\textwidth}{!}{
    \begin{tabular}{c|ccccc|ccc}
    \toprule
    \multirow{2}{*}{\textbf{System}} & \multicolumn{5}{c|}{\textbf{Automatic Evaluation}} & \multicolumn{3}{c}{\textbf{Human Evaluation}} \\
    {} & {R-1} & {R-2} & {R-L} & {B} & {D} & {FAC} & {COH} & {REL} \\
    \midrule
    \multicolumn{9}{c}{CNNDM} \\
    \midrule
    CRL-COM (R) & \textbf{45.75} & \textbf{21.87} & \textbf{42.27} & -1.43 & 36.28 & 0.76 & 4.00 & \textbf{4.17} \\
    CRL-COM (B) & 41.07 & 18.15 & 36.63 & \textbf{-0.78} & 88.92 & \textbf{0.99} & \textbf{4.05} & 3.96 \\
    CRL-COM (D) & 42.20 & 19.21 & 38.19 & -0.80 & \textbf{89.48} & \textbf{0.99} & 4.03 & 4.04 \\
    \midrule
    \multicolumn{9}{c}{XSUM} \\
    \midrule
    CRL-COM (R) & \textbf{47.28} & \textbf{24.14} & \textbf{38.78} & -2.42 & 32.75 & 0.38 & 3.52 & 3.25 \\
    CRL-COM (B) & 41.85 & 19.38 & 33.46 & \textbf{-1.87} & 37.48 & \textbf{0.51} & \textbf{3.73} & \textbf{3.50} \\
    CRL-COM (D) & 44.38 & 22.16 & 36.57 & -2.38 & \textbf{40.91} & 0.50 & 3.62 & 3.29 \\
    \bottomrule
    \end{tabular}
}
\caption{Results of each system on CNNDM and XSUM dataset. Note that R stands for ROUGE, B stands for BARTScore, and D stands for DAE.}
\label{tab:fewshot1000}
\end{table*}

\subsection{Results and Analysis}
\paragraph{Contrastive reward learning can enforce models to learn from feedback of factuality metrics}
Driven by \textbf{Q1}, we observe that results from automatic evaluation presented in \autoref{tab:fewshot1000} indicate that contrastive reward learning enables abstractive summarization models to develop in a direction that aligns with existing factuality metrics.

\paragraph{Learning from factuality metrics improves factuality of abstractive summarization.}
Driven by \textbf{Q2}, we observe that results from human evaluation presented in \autoref{tab:fewshot1000} indicate that on both datasets, CRL-COM (B) and CRL-COM (D) exhibit superior performance in terms of factuality compared to CRL-COM (R). This suggests that while learning from factuality metrics such as BARTScore and DAE may potentially result in sacrificing the performance of the models on ROUGE scores, the resulting models can generate more factually consistent summaries. In other words, summaries with higher BARTScore or DAE scores but lower ROUGE scores tend to be more factually consistent with the source article compared to those with lower BARTScore or DAE scores but higher ROUGE scores. This further supports the assertion that BARTScore and DAE are effective at capturing factual information.

\paragraph{Learning from factuality metrics did not sacrifice coherence and relevance.}
According to human evaluations, the summaries generated by CRL-COM (B) and CRL-COM (D) showed comparable coherence and relevance to those generated by CRL-COM (R). This suggests that BARTScore and DAE has comparable abilities to ROUGE in terms of measuring coherence and relevance.

\section{Related Work}

\subsection{Factuality Metrics for Abstractive Summarization}
Various factuality metrics assess the factual consistency between a summary and its corresponding source document. QA-based factuality metrics leverage question generation (QG) models to generate questions from the summary and question answering (QA) models to answer those questions, given both the source and summary \cite{wang-etal-2020-asking, durmus-etal-2020-feqa, scialom-etal-2021-questeval, fabbri-etal-2022-qafacteval}. Factuality is then evaluated based on the alignment between the answers from the source and summary. Another class of metrics, entailment-based factuality metrics \citep{kryscinski-etal-2020-evaluating, goyal-durrett-2021-annotating, laban-etal-2022-summac}, evaluates whether all the information in the summary is entailed by the source document. Recent studies on leveraging pre-trained language model as evaluation \citep{yuan2021bartscore} also achieve competitive performance on evaluating factuality.

\subsection{Improving Factuality of Abstractive Summarization via Contrastive Learning}

Several contrastive learning frameworks have been proposed to enable models to learn factuality from positive samples (such as reference summaries) and negative samples (such as edited reference summaries and system generated summaries). For example, CLIFF \citep{cao-wang-2021-cliff} and CO2Sum \citep{liu2021co2sum}. Both of which are similar in nature but CO2Sum employs more sophisticated methods for negative sample construction. 

\section{Conclusion}
In this work, we present a simple contrastive reward learning framework that enforces abstractive summarization models to learn from feedback of existing factuality metrics. Empirical studies demonstrate the effectiveness of this approach, showing that abstractive summarization models that learn from factuality metric feedback through contrastive reward learning can generate more factual summaries without sacrificing coherence or relevance. This suggests that further advancements in the reward learning paradigm and factuality metrics can facilitate the development of more factually consistent abstractive summarization models.

\section{Limitations}
While we have included two distinctive dataset (CNNDM and XSUM) in our experiments, more non-news datasets could be included in future studies. Other possibilities for future work include comparing the capability of RL-based reward learning and contrastive reward learning in improving the factuality of abstractive summarization models.

\section{Ethics Statement}
Even though some of the investigated systems may achieve a high level of factuality on the CNNDM dataset, this does not guarantee that they can be used as off-the-shelf factual consistent summarization models. Thorough evaluation should be conducted before using these models in high-stakes settings to ensure their reliability.

\section*{Acknowledgements}
We would like to thank Yixin Liu for helpful discussion on BRIO. We would also like to thank Tanya Goyal for helpful discussion on DAE.


\bibliography{anthology,custom}
\bibliographystyle{acl_natbib}

\appendix


\end{document}